\documentclass[letterpaper, 10 pt, conference]{ieeeconf}  

\usepackage{booktabs}
\usepackage{graphicx} 
\usepackage{url}
\usepackage{hyperref}
\usepackage{caption}
\usepackage{marvosym}  
\usepackage{amsmath,amssymb}
\usepackage{algorithm}
\usepackage{algpseudocode}
\usepackage{url}
\IEEEoverridecommandlockouts                              

\title{\LARGE \bf
Scene-Conditioned Relation Routing for urban cellular activity forecasting
}

\author{Qingzhong Li$^{1}$, Jingye Lin$^{1}$, Hui Ma$^{1,*}$, Yajun Zhang$^{1}$, Xinjun Pei$^{1}$, Ming Yan$^{1}$, and Fei Xing$^{2}$
\thanks{*Corresponding author: Hui Ma
        (e-mail: huima@xju.edu.cn)}%
\thanks{$^{1}$Qingzhong Li, Jingye Lin, Hui Ma, Yajun Zhang, Xinjun Pei, and Ming Yan are with Xinjiang Key Laboratory of Intelligent Computing and Smart Applications,
        School of Software, Xinjiang University, Urumqi 830017, China.}%
\thanks{$^{2}$Fei Xing is with the College of Geography and Remote Sensing Sciences,
        Xinjiang University, Urumqi 830046, China.}%
}

\begin{document}

\maketitle

\thispagestyle{empty}
\pagestyle{empty}

\begin{abstract}

Urban cellular activity forecasting requires jointly modeling heterogeneous spatiotemporal signals, including SMS usage, mobile network traffic, and call activity. Existing methods often separate temporal modeling, spatial relation learning, and multi-signal prediction, relying on fixed graph structures or static multi-task learning schemes, which limits their adaptability to changing urban scenes. We propose SCRR-Net, a scene-conditioned spatial relation routing framework in which urban contextual information jointly controls spatial dependency selection and cross-task knowledge transfer. SCRR-Net includes a context encoder, a spatial graph expert routing module, a temporal Transformer encoder, and a task knowledge routing module. Experiments on the Milano and Trento datasets demonstrate that SCRR-Net consistently outperforms competing methods on SMS, network traffic, and call activity forecasting, while providing interpretable routing behaviors.

\end{abstract}

\section{Introduction}

Urban cellular systems generate large volumes of spatiotemporal signals that reflect collective human activity across cities, including Short Message Service (SMS) usage, mobile network traffic, and call activity. Forecasting these signals is important for network resource allocation, anomaly detection, and intelligent urban infrastructure management~\cite{Chai2025UoMo,Li2025HSLTNet}. In practice, however, these cellular signals are heterogeneous in both temporal dynamics and statistical characteristics. SMS activity is often bursty and event-sensitive, mobile network traffic usually exhibits smoother yet highly periodic patterns, and call activity follows its own behavioral rhythm. As a result, accurate forecasting requires jointly modeling heterogeneous signal modalities together with their temporal evolution. Meanwhile, spatial dependencies among urban regions also play a critical role, since activity propagation and synchronization are often shaped by geographic proximity, functional similarity, and scene-dependent cross-region interactions~\cite{Hussien2025Spatiotemporal5G}.

Despite recent progress in urban cellular activity forecasting, existing methods still face substantial challenges in modeling scene-dependent spatiotemporal dependencies. Sequence-based models, such as recurrent networks and Transformers\cite{PatchTST}, mainly focus on temporal representation learning, but they usually lack explicit mechanisms to capture spatial interactions among urban regions, which limits their ability to model coupled regional dynamics. Graph-based spatiotemporal models\cite{PDFormer,STAEformer} alleviate this issue by introducing relational structures, yet many of them rely on a single predefined graph and therefore cannot adapt to changing interaction patterns under different urban scenes. Although adaptive graph learning methods\cite{TRL-Trans} provide greater flexibility by dynamically estimating adjacency structures, unconstrained graph construction may introduce instability , especially when relational signals are weak or highly variable.

Another important challenge lies in multi-signal forecasting, where SMS, Net, and Call signals exhibit both shared urban regularities and signal-specific behaviors. Existing multi-task learning methods typically improve prediction by sharing representations across related tasks, but most of them adopt static parameter sharing or feature-driven expert allocation~\cite{Zhou2023HiNet,Zhang2025ODMTL}, which cannot explicitly determine when different signals should share knowledge and when task-specific modeling should dominate. As a result, these methods are often unable to adapt their knowledge-sharing patterns to changing urban scenes, making them vulnerable to negative transfer across heterogeneous cellular signals. Therefore, a desirable forecasting framework should jointly adapt spatial dependency modeling and cross-task knowledge transfer according to the current urban context.

To address these limitations, we reformulate urban cellular activity forecasting as a \emph{scene-conditioned relation routing} problem, where urban context jointly governs spatial dependency selection and cross-task knowledge transfer. Different urban scenes, such as commuting hours, nighttime periods, holidays, and unusual events, may induce different coupling patterns among regions and different sharing behaviors across SMS, Net, and Call signals. Based on this observation, we propose \textbf{SCRR-Net}, a Scene-Conditioned Spatial Relation Routing Network. SCRR-Net uses a context encoder to derive a latent urban-state representation, which serves as a unified routing signal for both spatial and task-level modeling. Specifically, it guides a spatial graph expert routing module to combine multiple spatial relation priors and controls a task knowledge routing module to balance shared and task-specific experts across signals. A temporal Transformer encoder is further introduced to capture long-range dependencies from routed region-level representations. In this way, SCRR-Net unifies temporal modeling, scene-aware spatial reasoning, and adaptive cross-task knowledge transfer within a single framework.


The main contributions of this work are summarized as follows:
\begin{itemize}
    \item We propose SCRR-Net, a unified scene-conditioned relation routing framework for urban cellular activity forecasting, which jointly models spatial dependency selection and cross-task knowledge transfer under changing urban scenes.
    \item We introduce a spatial graph expert routing mechanism that adaptively combines multiple complementary spatial relation priors, allowing the model to capture scene-dependent spatial interactions while maintaining structural stability and interpretability.
    \item We design a task knowledge routing architecture with shared and task-specific experts for multi-signal forecasting, which improves cross-signal knowledge transfer and alleviates negative transfer among heterogeneous cellular signals.
\end{itemize}

\section{RELATED WORK}

\subsection{Temporal Dependency Modeling}

Early studies on cellular traffic forecasting mainly focused on learning temporal dependencies from historical observations, aiming to capture short-term fluctuations, long-range periodicity, and recurring temporal patterns. For example, Shen \emph{et al.}~\cite{Shen2021TWACNet} proposed TWACNet, which enhances long-range temporal modeling through a time-wise attention mechanism. More recent works further improved temporal representation learning. For instance, Ma \emph{et al.}~\cite{Ma2025DSSMAttention} employed deep state space models with attention to better capture traffic dynamics over time. Although these methods improve sequential modeling, temporal-only approaches remain insufficient for cellular traffic forecasting because traffic evolution is also strongly influenced by spatial interactions among neighboring cells or urban regions.
\begin{figure}[t]
  \centering
  \includegraphics[width=\columnwidth]{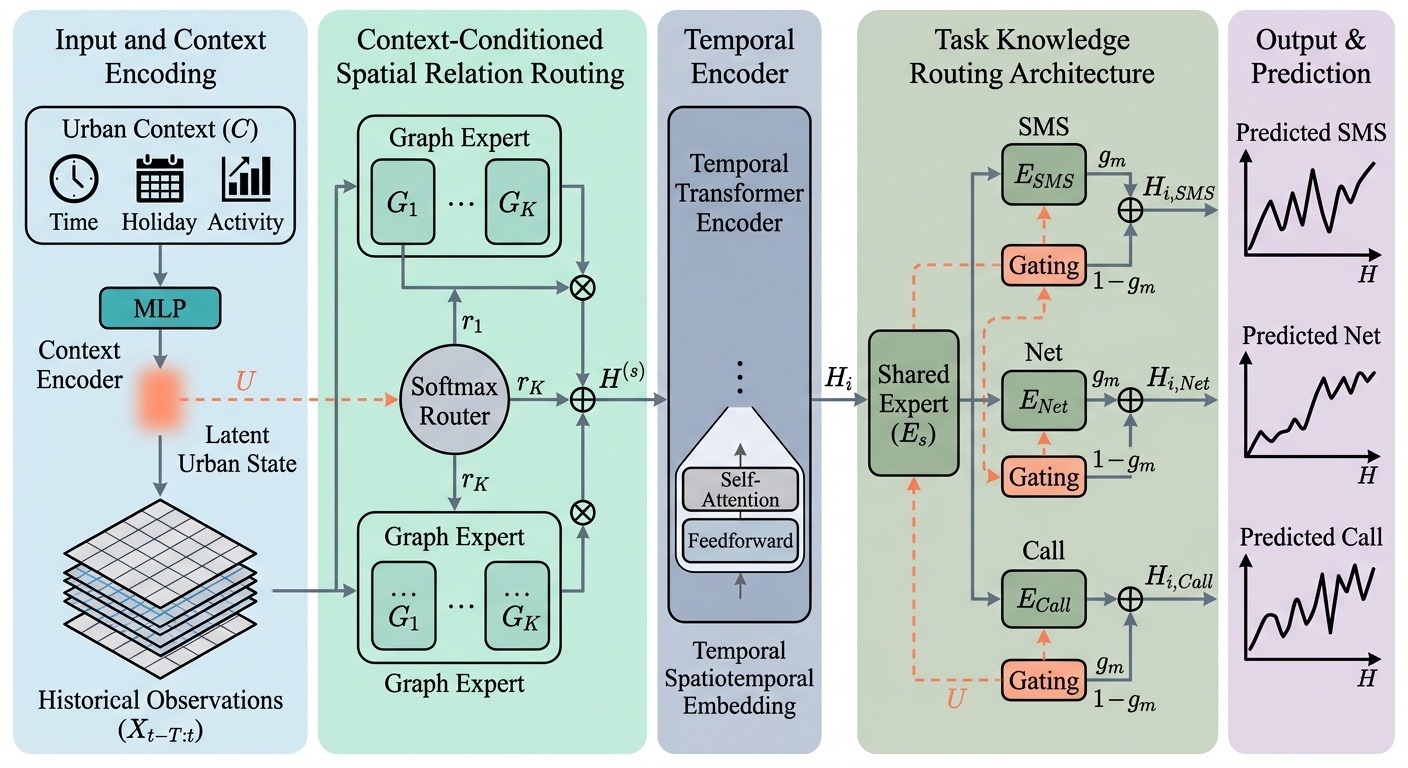}
  \caption{Overview of SCRR-Net for multi-signal urban communication forecasting. The framework consists of four stages: scene encoding, spatial graph expert routing, temporal representation learning, and task knowledge routing.}
  \label{fig:framework_overview}
\end{figure}
\subsection{Spatial Dependency Modeling}

To characterize spatial correlations in communication networks, subsequent studies extended temporal forecasting to spatiotemporal modeling. Weng \emph{et al.}~\cite{Weng2023DDGCRN} proposed DDGCRN, which introduces signal decomposition into a dynamic graph convolutional recurrent framework to model complex spatiotemporal dependencies. Li \emph{et al.}~\cite{Li2026GraFSTNet} developed GraFSTNet, which combines graph-based spatial learning with time-frequency analysis to better characterize periodic traffic patterns. In addition, Wang \emph{et al.}~\cite{Wang2023AHSTGNN} proposed AHSTGNN to learn adaptive hybrid spatial relations while considering multi-periodic temporal inputs and cell-level heterogeneity, whereas Zhou \emph{et al.}\cite{TRL-Trans} proposed TRL-Trans, which incorporates adaptive graph learning into temporal representation learning to capture dynamic spatiotemporal characteristics. However, existing methods still either rely on a single relational prior or learn graph structures in a largely unconstrained manner, making it difficult to explicitly select and combine multiple complementary spatial relation patterns under different urban scenes.

\subsection{Multi-Task Learning and Expert Routing Mechanisms}

Multi-task learning has been widely adopted to improve forecasting performance by exploiting shared information across related signals or tasks. In telecommunication forecasting, Gou and Zhang~\cite{Gou2023TelecommunicationMTL} proposed a multi-task framework that jointly models dependencies across different telecom signals. Hermosillo-Reynoso \emph{et al.}~\cite{HermosilloReynoso2025VTS} proposed a Transformer-based multi-task learning framework that employs shared parameters together with task-specific output heads to jointly model multiple vehicle traffic surveillance tasks. More recently, expert-based architectures have been introduced to improve modeling flexibility and specialization. For example, Lee and Ko~\cite{Lee2024TESTAM} proposed TESTAM, which introduces a mixture-of-experts mechanism to model diverse temporal and spatiotemporal traffic patterns, while Liu \emph{et al.}~\cite{Liu2026UrbanMoE} developed UrbanMoE, a sparse multi-modal mixture-of-experts framework for urban multi-task learning. However, most existing methods mainly focus on shared/task-specific decomposition or expert allocation itself. They usually rely on static parameter sharing or input-driven routing, without explicitly modeling when task knowledge should be shared under different urban scenes. In addition, task-level knowledge transfer is often designed independently of spatial relation modeling, leaving their interaction insufficiently explored.

%

Overall, existing studies have advanced temporal modeling, spatial relation learning, and cross-task knowledge sharing, but they still lack a unified context-conditioned mechanism that jointly modulates spatial relation selection and cross-task transfer.

\section{Method}
\label{sec:method}

We formulate multi-signal urban communication forecasting as a \emph{scene-conditioned relation routing} problem. Given a historical observation window and the aligned urban context, SCRR-Net first encodes the current scene into a latent urban-state representation, and then uses this scene representation to route spatial relation experts and regulate cross-task knowledge transfer. As illustrated in Fig.~\ref{fig:framework_overview}, the overall pipeline contains four stages: scene encoding, spatial graph expert routing, temporal representation learning, and task knowledge routing for multi-signal prediction.

\subsection{Problem Formulation}
\label{subsec:problem_formulation}

Let the city be partitioned into $N$ spatial regions, and let $M$ denote the number of communication signal types. In our setting, $M=3$, corresponding to SMS, Net, and Call. At each time step $\tau$, the communication observations over all regions are denoted by $X^{(\tau)} \in \mathbb{R}^{N \times M}$, where the $i$-th row $x_i^{(\tau)} \in \mathbb{R}^{M}$ represents the multi-signal communication activity of region $i$ at time $\tau$.

Given a historical observation window of length $T$, the model input consists of the past observation sequence $\mathcal{X}_t = \{X^{(t-T+1)}, X^{(t-T+2)}, \dots, X^{(t)}\}$ and an aligned contextual vector $c_t \in \mathbb{R}^{d_c}$ associated with the current forecasting window, where $d_c$ is the context feature dimension. The context vector may include temporal indicators such as time-of-day and day-of-week, as well as holiday indicators and other city-level auxiliary statistics. The forecasting objective is to predict the communication signals in the next $H$ time steps over all regions and all signal types. Formally, SCRR-Net learns the mapping
\begin{equation}
\hat{\mathcal{Y}}_t = f(\mathcal{X}_t, c_t),
\label{eq:forecast_mapping}
\end{equation}
where $\hat{\mathcal{Y}}_t \in \mathbb{R}^{N \times H \times M}$ denotes the predicted future communication sequence.

\subsection{Scene Encoding}
\label{subsec:scene_encoding}

To obtain a compact representation of the current urban scene, we encode the contextual vector $c_t$ into a latent urban-state embedding $u_t \in \mathbb{R}^{d_u}$. Specifically, the scene encoder is implemented as a two-layer multilayer perceptron:
\begin{equation}
u_t = \phi(c_t) = W_2\,\mathrm{ReLU}(W_1 c_t + b_1) + b_2.
\label{eq:scene_encoder}
\end{equation}

The latent vector $u_t$ serves as a scene controller. Instead of directly concatenating contextual features with node observations, SCRR-Net uses $u_t$ to govern routing decisions in both the spatial branch and the task branch. In this way, the model explicitly separates scene interpretation from downstream forecasting, which makes the routing behavior easier to interpret.

\subsection{Scene-Conditioned Spatial Relation Routing}
\label{subsec:spatial_routing}

\begin{figure}[t]
  \centering
  \includegraphics[width=\columnwidth]{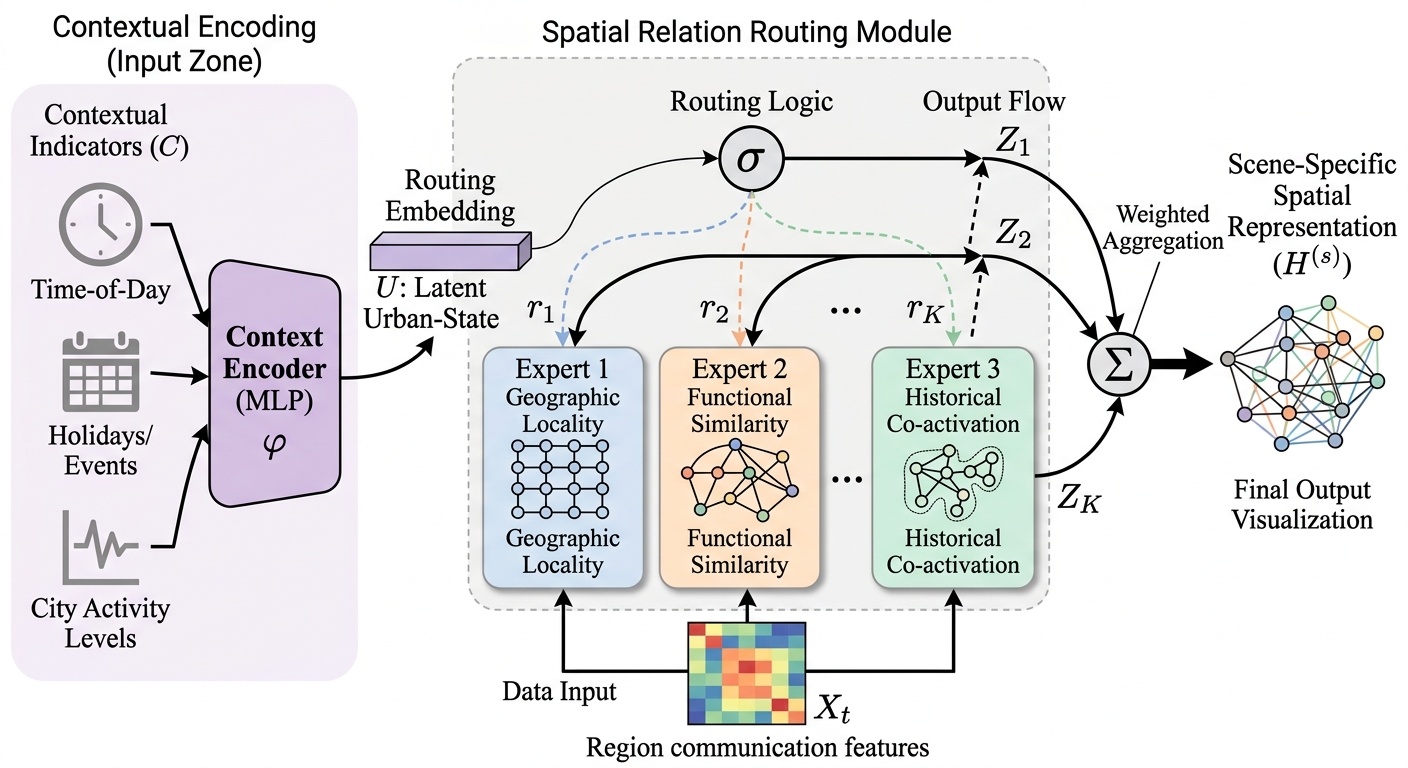}
  \caption{Scene-conditioned routing among spatial graph experts representing different urban relation priors. For illustration, three representative graph experts are shown, while the formulation is defined for a general number $K$ of experts.}
  \label{fig:graph_routing}
\end{figure}

To model scene-dependent spatial interactions, we predefine $K$ spatial graph experts $\{G^{(1)}, G^{(2)}, \dots, G^{(K)}\}$. Each expert is associated with an adjacency matrix $A^{(k)} \in \mathbb{R}^{N \times N}$ that encodes one type of spatial relation prior among regions. These priors may capture complementary dependencies such as geographic locality, historical co-activation, and functional similarity. In our implementation, we instantiate three representative spatial priors consistent with Fig.~\ref{fig:graph_routing}. The first expert is a geographic locality graph, where nearby regions are connected according to spatial distance. Let $d_{ij}$ denote the geographic distance between regions $i$ and $j$. Its edge weight is defined as
\begin{equation}
A^{(\mathrm{geo})}_{ij} =
\begin{cases}
\exp\!\left(-d_{ij}^2 / \sigma_d^2\right), & d_{ij} \le \delta,\\
0, & \text{otherwise},
\end{cases}
\label{eq:geo_graph}
\end{equation}
where $\sigma_d$ controls the distance decay and $\delta$ is a distance threshold.

The second expert is a functional similarity graph, which captures long-term similarity between regional communication profiles. Let $p_i \in \mathbb{R}^{d_f}$ denote the long-term activity descriptor of region $i$. Its edge weight is computed by cosine similarity:
\begin{equation}
A^{(\mathrm{fun})}_{ij}
=
\max\!\left(
0,\;
\frac{p_i^\top p_j}{\|p_i\|_2 \|p_j\|_2}
\right).
\label{eq:functional_graph}
\end{equation}

The third expert is a historical co-activation graph, which reflects synchronization between regions in past communication dynamics. Let $v_i$ and $v_j$ denote the historical multi-signal activity sequences of regions $i$ and $j$, respectively. We define
\begin{equation}
A^{(\mathrm{coa})}_{ij}
=
\max\!\left(0,\; \mathrm{Corr}(v_i, v_j)\right),
\label{eq:coactivation_graph}
\end{equation}
where $\mathrm{Corr}(\cdot,\cdot)$ denotes the Pearson correlation coefficient.

For each prior graph, we retain the top-$r$ neighbors for each region, symmetrize the adjacency matrix, and set diagonal entries to zero before applying the normalization in \eqref{eq:adj_norm}. In this way, SCRR-Net routes among multiple stable spatial priors according to the inferred urban scene, rather than relying on a single fixed graph.

We first normalize each adjacency matrix as
\begin{equation}
\tilde{A}^{(k)} = D_k^{-\frac{1}{2}} (A^{(k)} + I) D_k^{-\frac{1}{2}},
\label{eq:adj_norm}
\end{equation}
where $D_k$ is the degree matrix of $A^{(k)} + I$.

For each historical step $\tau \in [t-T+1, t]$, graph expert $k$ performs spatial propagation on the communication signal matrix:
\begin{equation}
Z_k^{(\tau)} = \sigma \left( \tilde{A}^{(k)} X^{(\tau)} W_k + b_k \right),
\label{eq:graph_expert}
\end{equation}
where $W_k \in \mathbb{R}^{M \times d_s}$ is a learnable projection matrix, $b_k$ is a bias term, and $\sigma(\cdot)$ denotes a nonlinear activation function.

The contribution of each graph expert is determined by the current scene embedding $u_t$. Specifically, the routing weight of expert $k$ is defined as
\begin{equation}
\alpha_{k,t}
=
\frac{\exp(q_k^\top u_t)}
{\sum_{j=1}^{K} \exp(q_j^\top u_t)},
\label{eq:spatial_routing_weight}
\end{equation}
where $q_k \in \mathbb{R}^{d_u}$ is a learnable routing vector associated with expert $k$.

The routed spatial representation at time step $\tau$ is then obtained by aggregating all expert outputs:
\begin{equation}
S^{(\tau)} = \sum_{k=1}^{K} \alpha_{k,t} Z_k^{(\tau)}, \qquad \tau = t-T+1, \dots, t.
\label{eq:spatial_fusion}
\end{equation}

Here, the same scene embedding $u_t$ is used to modulate all historical steps within the current forecasting window. This design matches the assumption that one forecasting sample corresponds to one scene state. Therefore, instead of learning a fully unconstrained dynamic graph from scratch, SCRR-Net adaptively composes multiple stable spatial relation priors according to the inferred urban scene.

\subsection{Temporal Representation Learning}
\label{subsec:temporal_learning}

After scene-conditioned spatial fusion, we obtain a routed spatial representation sequence over the historical window. For each region $i$, we extract its temporal sequence from the routed spatial features, written as $\mathbf{s}_{i,t} = [s_i^{(t-T+1)}, s_i^{(t-T+2)}, \dots, s_i^{(t)}] \in \mathbb{R}^{T \times d_s}$, where $s_i^{(\tau)}$ denotes the $i$-th row of $S^{(\tau)}$.

We then feed this regional sequence into a temporal Transformer encoder to capture long-range temporal dependencies:
\begin{equation}
\mathbf{H}_{i,t} = \mathrm{TransEnc}(\mathbf{s}_{i,t} + P),
\label{eq:temporal_encoder}
\end{equation}
where $P \in \mathbb{R}^{T \times d_s}$ denotes positional embeddings. The final regional spatiotemporal representation is obtained through a pooling operation:
\begin{equation}
h_{i,t} = \mathrm{Pool}(\mathbf{H}_{i,t}) \in \mathbb{R}^{d_h}.
\label{eq:regional_representation}
\end{equation}

The resulting vector $h_{i,t}$ summarizes the routed spatial interactions and temporal evolution of region $i$ over the historical window, and it serves as the input to the task knowledge routing module.

\subsection{Task Knowledge Routing and Multi-Signal Prediction}
\label{subsec:task_routing}

\begin{figure}[t]
   \centering
  \includegraphics[width=\columnwidth]{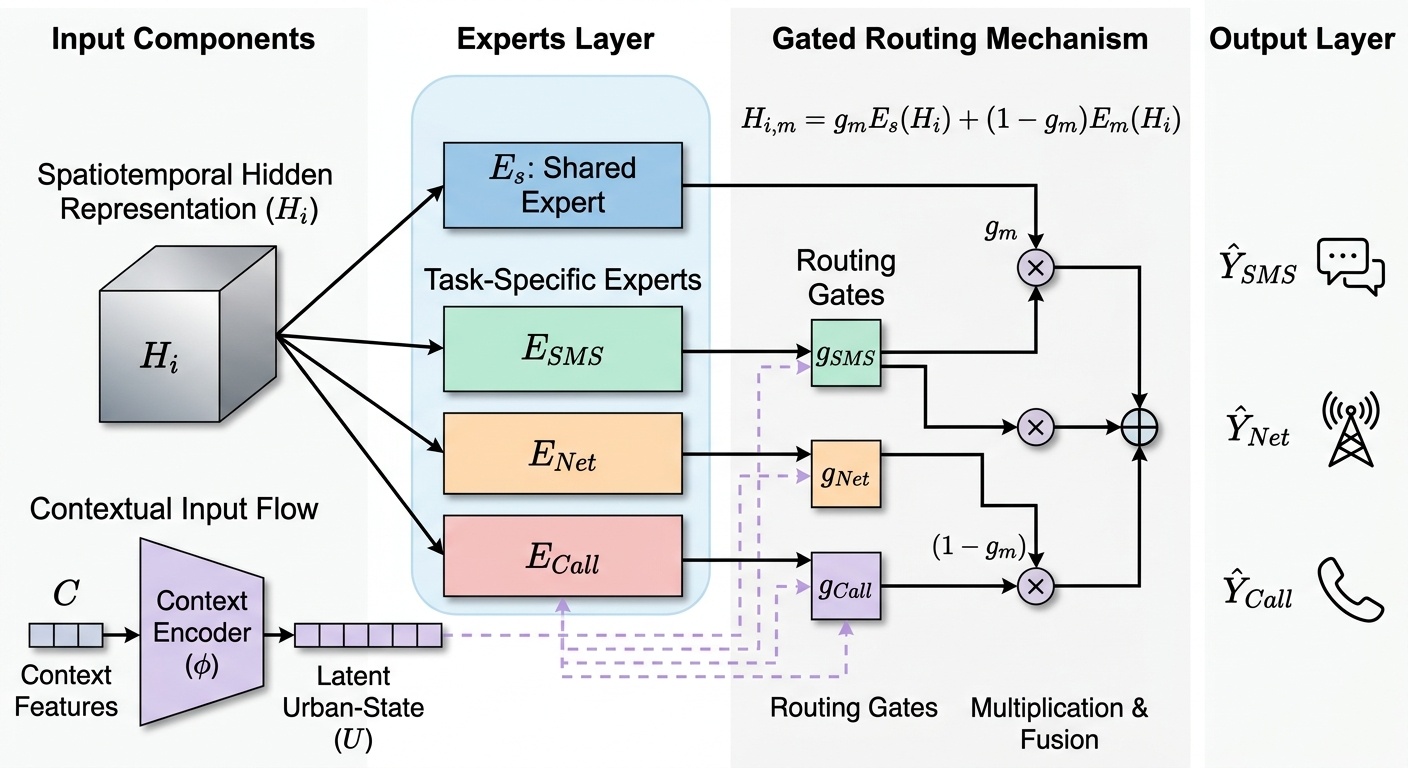}
   \caption{Task knowledge routing with shared and task-specific experts for multi-signal forecasting.}
   \label{fig:task_routing}
\end{figure}

Different communication signals share common urban driving factors, but they also exhibit distinct statistical properties. To balance shared knowledge and signal-specific modeling, we adopt a task knowledge routing architecture consisting of one shared expert and $M$ task-specific experts, as shown in Fig.~\ref{fig:task_routing}.

The shared expert captures common communication dynamics and is defined as
\begin{equation}
e^{\mathrm{sh}}_{i,t} = E_{\mathrm{sh}}(h_{i,t}),
\label{eq:shared_expert}
\end{equation}
where $E_{\mathrm{sh}}(\cdot)$ is implemented as a two-layer multilayer perceptron. For each signal type $m \in \{1, \dots, M\}$, a task-specific expert produces
\begin{equation}
e^{\mathrm{pr}}_{i,m,t} = E_m(h_{i,t}),
\label{eq:private_expert}
\end{equation}
where $E_m(\cdot)$ focuses on the signal-dependent patterns of task $m$.

The routing gate for task $m$ is generated from the same scene embedding $u_t$:
\begin{equation}
g_{m,t} = \sigma(a_m^\top u_t + b_m),
\label{eq:task_gate}
\end{equation}
where $a_m \in \mathbb{R}^{d_u}$ and $b_m$ are learnable parameters.

The task-specific fused representation is then computed as
\begin{equation}
r_{i,m,t}
=
g_{m,t} \, e^{\mathrm{sh}}_{i,t}
+
(1-g_{m,t}) \, e^{\mathrm{pr}}_{i,m,t}.
\label{eq:task_fusion}
\end{equation}

This formulation means that the inferred urban scene directly controls how much each prediction task relies on shared knowledge versus task-specific knowledge. For routine scenes, the model may place more emphasis on the shared expert; for unusual scenes, it may rely more heavily on the task-specific branch.

Finally, each task uses its own prediction head to generate the future $H$-step forecast:
\begin{equation}
\hat{y}_{i,m,t}^{(1:H)} = \Psi_m(r_{i,m,t}) \in \mathbb{R}^{H},
\label{eq:prediction_head}
\end{equation}
where $\Psi_m(\cdot)$ is a linear layer or a lightweight multilayer perceptron. By stacking predictions over all regions and all signal types, we obtain the final output tensor $\hat{\mathcal{Y}}_t$.

\begin{table*}[t]
\caption{Main forecasting results on the Milano and Trento datasets.}
\label{tab:main_results}
\centering
\small
\setlength{\tabcolsep}{5pt}
\begin{tabular}{lcccccccccccc}
\toprule
{\textbf{Model}} 
& \multicolumn{6}{c}{\textbf{Milano}} 
& \multicolumn{6}{c}{\textbf{Trento}} \\
\cmidrule(lr){2-7} \cmidrule(lr){8-13}
& \multicolumn{2}{c}{\textbf{Call}} 
& \multicolumn{2}{c}{\textbf{SMS}} 
& \multicolumn{2}{c}{\textbf{Net}}
& \multicolumn{2}{c}{\textbf{Call}} 
& \multicolumn{2}{c}{\textbf{SMS}} 
& \multicolumn{2}{c}{\textbf{Net}} \\
\cmidrule(lr){2-3} \cmidrule(lr){4-5} \cmidrule(lr){6-7}
\cmidrule(lr){8-9} \cmidrule(lr){10-11} \cmidrule(lr){12-13}
& \textbf{MAE} & \textbf{RMSE} 
& \textbf{MAE} & \textbf{RMSE} 
& \textbf{MAE} & \textbf{RMSE}
& \textbf{MAE} & \textbf{RMSE} 
& \textbf{MAE} & \textbf{RMSE} 
& \textbf{MAE} & \textbf{RMSE} \\
\midrule
Transformer & 0.8661 & 1.4782 & 1.5357 & 3.4106 & 4.6668 & 8.6496 & 1.1414 & 1.9584 & 3.3291 & 6.6050 & 6.7552 & 10.3996 \\
PatchTST & 0.7910 & 1.3519 & 1.4219 & 3.0135 & 3.2163 & 5.7408 & 1.0458 & 1.8109 & 2.9409 & 5.9359 & 5.8640 & 9.2135 \\
LightGTS & 0.7354 & 1.2841 & 1.3520 & 2.8826 & 2.9257 & 5.4559 & 0.9837 & 1.7405 & 2.7768 & 5.6580 & 5.6179 & 8.8386 \\
STGCN & 0.6732 & 1.2047 & 1.2266 & 2.7204 & 2.7793 & 5.0881 & 0.9255 & 1.6584 & 2.6721 & 5.4535 & 5.4033 & 8.4050 \\
PDFormer & 0.5670 & 1.0098 & 1.0174 & 2.2999 & 2.3797 & 4.4034 & 0.7639 & 1.3759 & 2.2410 & 4.6554 & 4.6221 & 7.3922 \\
STAEformer & 0.5461 & 0.9773 & 0.9837 & 2.2383 & 2.3014 & 4.2507 & 0.7375 & 1.3361 & 2.1589 & 4.5135 & 4.4822 & 7.1770 \\
Graph WaveNet & 0.5121 & 0.9197 & 0.9233 & 2.1121 & 2.0977 & 3.8791 & 0.6876 & 1.2547 & 2.0141 & 4.2310 & 4.2324 & 6.7138 \\
MTGNN & 0.4874 & 0.8789 & 0.8717 & 2.0037 & 1.9622 & 3.7302 & 0.6470 & 1.1930 & 1.8861 & 4.0073 & 3.9655 & 6.3962 \\
MegaCRN & 0.4665 & 0.8485 & 0.8310 & 1.9257 & 1.8271 & 3.5549 & 0.6035 & 1.1334 & 1.7985 & 3.8365 & 3.7657 & 6.0645 \\
TRL-Trans & 0.4444 & 0.8098 & 0.7920 & 1.8369 & 1.7121 & 3.3415 & 0.5624 & 1.0637 & 1.6931 & 3.6119 & 3.5478 & 5.7366 \\
MMoE & 0.5749 & 1.0423 & 1.0746 & 2.3534 & 2.4892 & 4.6176 & 0.8021 & 1.4373 & 2.3665 & 4.8582 & 4.8189 & 7.5888 \\
PLE & 0.5303 & 0.9518 & 0.9603 & 2.1492 & 2.1795 & 4.0439 & 0.7170 & 1.2939 & 2.0725 & 4.3082 & 4.3842 & 6.9011 \\
HiNet & 0.4995 & 0.9022 & 0.8988 & 2.0534 & 2.0458 & 3.7724 & 0.6669 & 1.2202 & 1.9398 & 4.1054 & 4.0775 & 6.5117 \\
\midrule
\textbf{SCRR-Net (Ours)} 
& \textbf{0.3171} & \textbf{0.4420} 
& \textbf{0.4470} & \textbf{0.8809} 
& \textbf{1.5447} & \textbf{2.2971} 
& \textbf{0.3147} & \textbf{0.5191} 
& \textbf{0.8257} & \textbf{1.7801} 
& \textbf{2.0813} & \textbf{3.6622} \\
\bottomrule
\end{tabular}
\end{table*}

\section{EXPERIMENTS}

\subsection{Datasets}
\textbf{Milano.}
The Milano dataset\footnote{OpenStreet: \url{http://www.openstreet.com/}} is a multimodal urban dataset collected in Milan from November 1, 2013 to January 1, 2014. The city is divided into 10{,}000 grids, each covering about $235 \times 235$ m$^2$. We use hourly SMS, Net, and Call records as prediction targets.

\textbf{Trento.}
The Trento dataset\footnotemark[1] was collected from the Province of Trentino during the same period. The study area contains 6,575 grids. We use hourly SMS, Net, and Call records as prediction targets.




\subsection{Overall Comparison}


To verify the effectiveness of the proposed method, Table~\ref{tab:main_results} compares SCRR-Net with four categories of representative baselines. Pure temporal models (Transformer, PatchTST and LightGTS) mainly learn temporal dependencies from historical sequences and can capture short-term fluctuations and periodic patterns, but they lack explicit modeling of spatial interactions among regions, resulting in relatively limited overall performance. Spatiotemporal models (STGCN, PDFormer and STAEformer) jointly capture temporal dynamics and spatial dependencies, and generally outperform pure temporal methods, indicating the importance of spatial information for this task. However, some of these methods still rely on predefined structures and thus are less flexible when facing constantly changing urban interaction patterns. Adaptive graph learning methods (Graph WaveNet, MTGNN, MegaCRN and TRL-Trans) further improve the results, suggesting that dynamic relation modeling helps better capture inter-region interactions that vary across scenes. Nevertheless, unconstrained graph learning may also introduce noise or unstable relations. Multi-task learning methods  (MMoE, PLE and HiNet) achieve relatively strong performance on some signals, showing that cross-signal knowledge sharing is beneficial. However, since their sharing mechanisms usually lack scene awareness, they are still vulnerable to negative transfer when the heterogeneity across signals becomes large.

\begin{figure}[t]
  \centering
  \includegraphics[width=0.9\columnwidth]{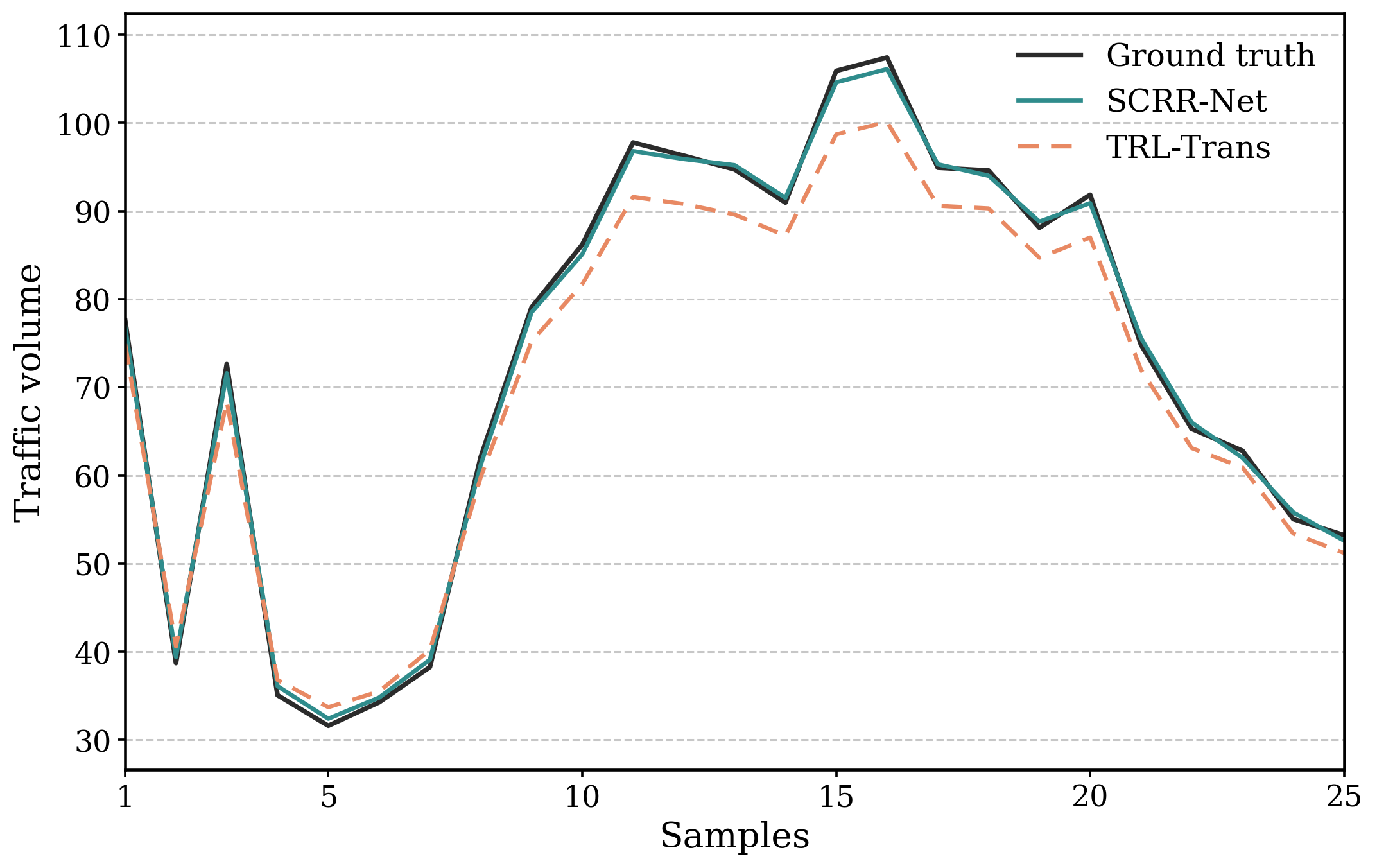}\hfill
  \includegraphics[width=0.9\columnwidth]{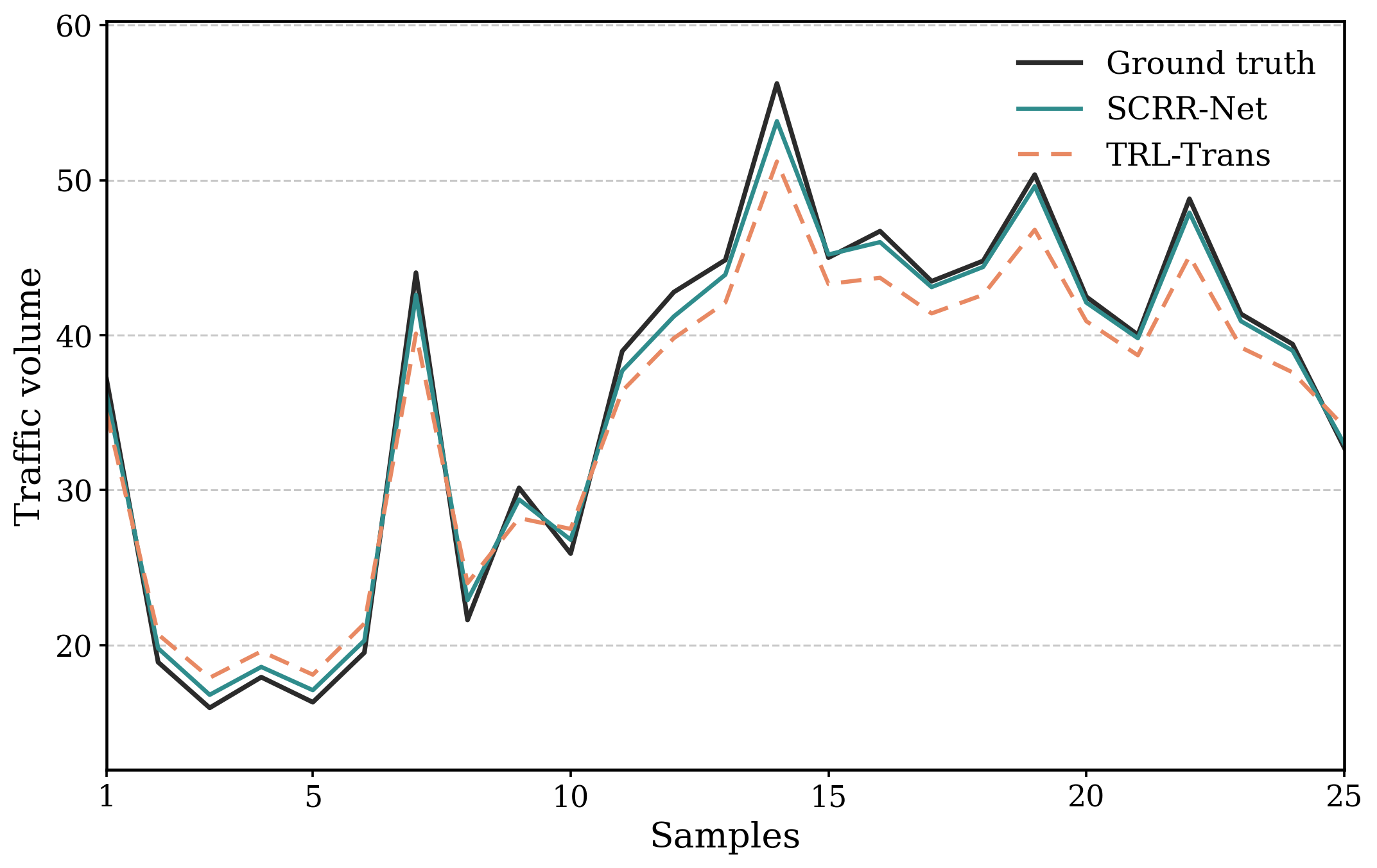}
  
  \caption{Comparison between the ground-truth values and the predicted values on the Milano ,Trento datasets.}
  \label{fig:prediction_case}
\end{figure}

In contrast, SCRR-Net achieves the best results on both datasets and across all three signal types. This demonstrates that the proposed method can not only leverage contextual information to adaptively select more suitable spatial relation priors, but also dynamically adjust the knowledge-sharing pattern across different signals according to the current urban scene, thereby simultaneously enhancing spatial modeling capability and cross-task transfer effectiveness. The experimental results validate the effectiveness of scene-conditioned spatial relation routing and task knowledge routing for multi-signal urban cellular activity forecasting. Figure~\ref{fig:prediction_case} illustrates the ground-truth and predicted Net traffic series on the Milano and Trento datasets. It can be observed that SCRR-Net tracks the underlying traffic dynamics more accurately and responds better to sudden variations than the second-best baseline.


\subsection{Ablation Study}
\begin{table*}[t]
\caption{Illustrative ablation layout on the Milano and Trento datasets.}
\label{tab:ablation_results}
\centering
\small
\setlength{\tabcolsep}{4.2pt}
\begin{tabular}{lcccccccccccc}
\toprule
{\textbf{Variant}} 
& \multicolumn{6}{c}{\textbf{Milano}} 
& \multicolumn{6}{c}{\textbf{Trento}} \\
\cmidrule(lr){2-7} \cmidrule(lr){8-13}
& \multicolumn{2}{c}{\textbf{Call}} 
& \multicolumn{2}{c}{\textbf{SMS}} 
& \multicolumn{2}{c}{\textbf{Net}}
& \multicolumn{2}{c}{\textbf{Call}} 
& \multicolumn{2}{c}{\textbf{SMS}} 
& \multicolumn{2}{c}{\textbf{Net}} \\
\cmidrule(lr){2-3} \cmidrule(lr){4-5} \cmidrule(lr){6-7}
\cmidrule(lr){8-9} \cmidrule(lr){10-11} \cmidrule(lr){12-13}
& \textbf{MAE} & \textbf{RMSE} 
& \textbf{MAE} & \textbf{RMSE} 
& \textbf{MAE} & \textbf{RMSE}
& \textbf{MAE} & \textbf{RMSE} 
& \textbf{MAE} & \textbf{RMSE} 
& \textbf{MAE} & \textbf{RMSE} \\
\midrule
\textbf{Full Model} 
& \textbf{0.3171} & \textbf{0.4420}
& \textbf{0.4470} & \textbf{0.8809}
& \textbf{1.5447} & \textbf{2.2971}
& \textbf{0.3147} & \textbf{0.5191}
& \textbf{0.8257} & \textbf{1.7801}
& \textbf{2.0813} & \textbf{3.6622} \\
w/o Context Encoder
& 0.3415 & 0.4718
& 0.4864 & 0.9417
& 1.6768 & 2.4816
& 0.3396 & 0.5564
& 0.8938 & 1.9175
& 2.2584 & 3.9517 \\
w/o Multi-Graph Routing
& 0.3728 & 0.5194
& 0.5348 & 1.0187
& 1.7926 & 2.6483
& 0.3816 & 0.6118
& 0.9724 & 2.0675
& 2.4681 & 4.2386 \\
w/o Task Experts
& 0.3462 & 0.4786
& 0.4941 & 0.9568
& 1.6924 & 2.5093
& 0.3448 & 0.5645
& 0.9071 & 1.9442
& 2.2875 & 3.9886 \\
w/o Sparse Routing
& 0.3294 & 0.4581
& 0.4668 & 0.9084
& 1.6085 & 2.3827
& 0.3268 & 0.5376
& 0.8563 & 1.8468
& 2.1567 & 3.7914 \\
\bottomrule
\end{tabular}
\end{table*}
Table~\ref{tab:ablation_results} presents the ablation results of the major components. Removing the context encoder causes the spatial graph expert routing to degenerate into a static form, preventing the model from adaptively switching spatial propagation patterns across different urban scenes and thus leading to a clear performance drop. Replacing multiple spatial graph experts with a single fixed graph compresses diverse spatial relation priors into one structure, making it difficult for the model to capture heterogeneous coupling patterns such as local diffusion, coordination among functionally related regions, and event-driven synchronization. Removing task knowledge routing forces the model to rely more heavily on shared representations, weakening its ability to handle signal heterogeneity and increasing the risk of negative transfer. Replacing sparse scene-conditioned routing with dense routing also degrades performance, suggesting that selective routing is more effective than uniformly mixing all experts. In contrast, the full model achieves the best performance by jointly balancing spatial dependency modeling and cross-task knowledge sharing through scene-conditioned spatial relation routing and task knowledge routing.

\section{CONCLUSION}

This paper formulates urban communication forecasting as a scene-conditioned relation routing problem, and proposes \textbf{SCRR-Net} to jointly drive spatial graph expert routing and task knowledge routing through a latent urban-state representation generated by a context encoder. Combined with a temporal Transformer encoder, SCRR-Net can adaptively coordinate spatial dependency modeling, cross-task knowledge transfer, and temporal dynamic learning under changing urban scenes. Experimental results show that SCRR-Net achieves superior performance in terms of MAE and RMSE, while also exhibiting interpretable changes in spatial relations and knowledge-sharing patterns. In future work, richer contextual perception and adaptive discovery of spatial graph experts could be explored to improve the applicability of the model in more complex urban forecasting scenarios.

\textbf{Acknowledgments.} 
This work was supported by the Xinjiang Key Industries Talent Support
Project (XJRC-2025-GX-ZDCY and XJRC-2025-ZZB-ZDXQ-005), the Natural
Science Foundation of Xinjiang Uygur Autonomous Region (2025D01C296),
the Autonomous Region Key R\&D Project under Grant 2025B04051, and the
Tianchi Talents--Young Doctor Program (5105250183m and 5105250183l).




\bibliographystyle{IEEEtran}   
\bibliography{mybibfile}      

\end{document}